\PassOptionsToPackage{dvipsnames, table, xcdraw}{xcolor}
\documentclass[twocolumn]{mc_nankai}

\usepackage[utf8]{inputenc} % allow utf-8 input
\usepackage[T1]{fontenc}    % use 8-bit T1 fonts
\usepackage[dvipsnames,table,xcdraw]{xcolor}
\usepackage{amsfonts}       % blackboard math symbols
\usepackage{nicefrac}       % compact symbols for 1/2, etc.
\usepackage{microtype}      % microtypography
\usepackage{wrapfig}

\usepackage{graphicx, subcaption}
\usepackage{multirow}
\usepackage{colortbl}
\usepackage{ulem}
\newcommand{\ourMthd}{ImageCritic}
\newcommand{\ourBench}{CriticBench}
\newcommand{\incarr}[2]{#1~{\ding{223}}~#2}
\definecolor{myblue}{rgb}{0.21,0.49,0.74}
\usepackage{pifont}
\newcommand{\listref}[1]{List.~\ref{#1}}
\newcommand{\cmark}{\ding{51}} % √
\newcommand{\xmark}{\ding{55}} % ✗
\title{The Consistency Critic: Correcting Inconsistencies in Generated Images \\ via Reference-Guided Attentive Alignment}

\author[1]{Ziheng Ouyang}
\author[2]{Yiren Song}
\author[3]{Yaoli Liu}
\author[1]{Shihao Zhu}
\author[1]{Qibin Hou$^{\dagger}$}
\author[1]{Ming-Ming Cheng}
\author[2]{Mike Zheng Shou}

\affiliation[1]{VCIP, Nankai University}
\affiliation[2]{Show Lab, National University of Singapore\\}
        \affiliation[3]{State Key Laboratory of CAD\&CG, Zhejiang University}

\affiliationenter[]{$^{\dagger}$Corresponding author.}

\abstract{
Previous works have explored various customized generation tasks given a reference image, but they still face limitations in generating consistent fine-grained details. 
In this paper, our aim is to solve the inconsistency problem of generated images by applying a reference-guided post-editing approach and present our~\ourMthd{}.
We first construct a dataset of reference-degraded-target triplets obtained via VLM-based selection and explicit degradation, which effectively simulates the common inaccuracies or inconsistencies observed in existing generation models.
Furthermore, building on a thorough examination of the model's attention mechanisms and intrinsic representations, we accordingly devise an attention alignment loss and a detail encoder to precisely rectify inconsistencies.
\ourMthd{} can be integrated into an agent framework to automatically detect inconsistencies and correct them with multi-round and local editing in complex scenarios.
Extensive experiments demonstrate that~\ourMthd{} can effectively resolve detail-related issues in various customized generation scenarios, providing significant improvements over existing methods.
}

\mclink[Project Page]{\url{https://ouyangziheng.github.io/ImageCritic-Page/}}

\begin{document}
\maketitle
\justifying
\section{Introduction}
\begin{figure*}
    \centering
    \includegraphics[width=1\linewidth]{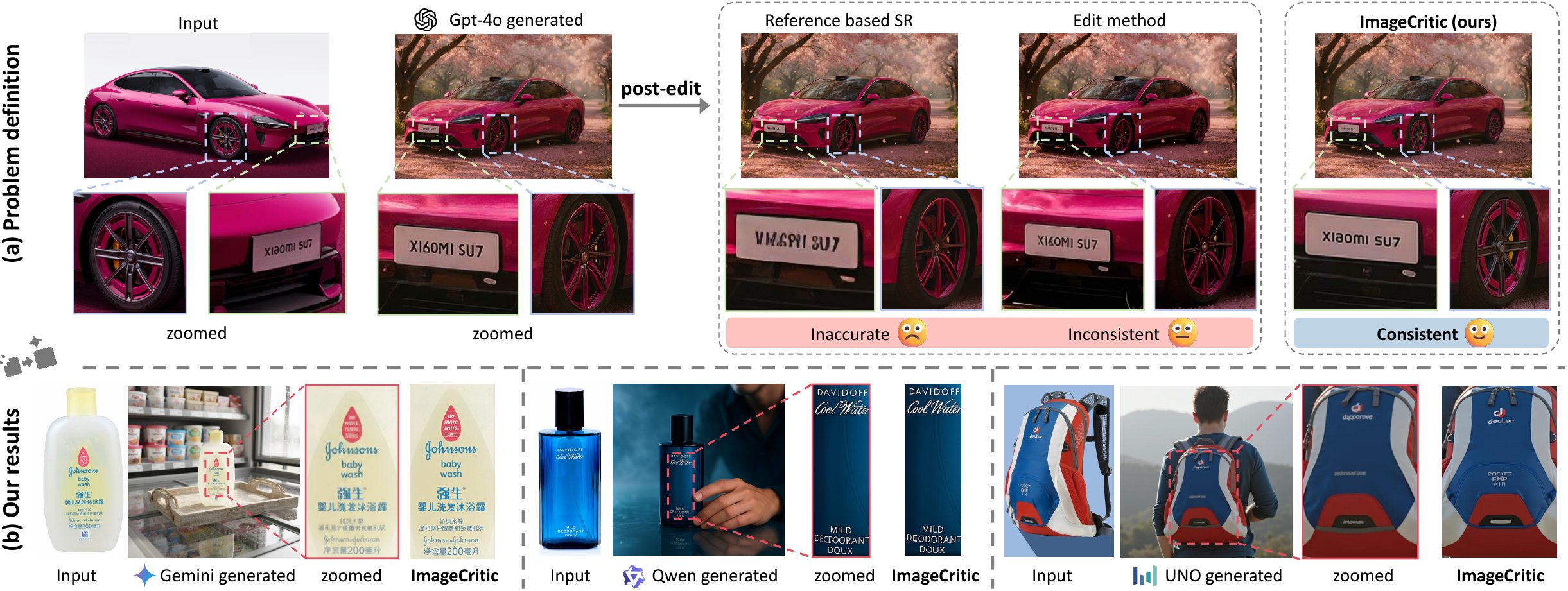}
    \caption{\textbf{Visual illustrations}. (a) illustrates that we first conduct customized generation using GPT-4o~\cite{gpt}, and then apply different methods for the post editing. Edit method~\cite{qwenimage} struggle to achieve fine-grained consistent generation, while images processed by super-resolution methods~\cite{guo2024refir} often exhibit noticeable detail inaccuracies. In contrast, our proposed \ourMthd{} corrects local details to ensure text and logo consistency while maintaining accurate spatial alignment, significantly improving the overall coherence of generated images. (b) We further apply our method to customized results generated by both state-of-the-art closed-source~\cite{nanobanana} and open-source models~\cite{qwenimage,uno}. After performing our correction, the fine-grained details of the generated images align precisely with those of the original objects, demonstrating the superior performance of our approach.
}
    \label{teaser} 
\end{figure*}
\label{sec:intro}
In recent years, image generation systems based on diffusion models have evolved from UNet architectures to Transformer-based Diffusion Transformer (DiT) models~\cite{zhang2021rstnet,esser2024scaling,flux}, with reference-guided generation becoming one of the key research trends, powering applications such as virtual try-on~\cite{ni2025itvton,meng2025hf,guo2025any2anytryon}, image editing~\cite{li2025visualcloze,han2024ace,xia2025dreamomni,xiao2025omnigen,xie2025anyrefill,huang2025photodoodle, yan2025eedit, feng2025dit4edit, wang2024cove, yan2025eedit}, 
and subject customization~\cite{chen2025unireal,ominicontrol,han2024ace,mosaic,xverse,uno,zhou2025dreamrenderer,ab,klora}. 
However, due to discrepancies introduced by VAE-based encoding and decoding in encoder–decoder frameworks and the loss of shallow-layer information in decoder-only structures, existing models often suffer from inconsistent and inaccurate fine-grained details between the generated image and the reference image.
As illustrated in~\figref{teaser}, images generated using current state-of-the-art~\cite{gpt,nanobanana,qwenimage,uno} customized generation methods commonly exhibit inconsistencies or blurriness in textual and logo regions.

One naive approach that aims to solve this inconsistency problem between generated images and reference images is reference-based super-resolution models, such as ReFIR~\cite{guo2024refir}, which refine blurry regions to improve alignment with the reference.
However, as shown in ~\figref{teaser}(a), the inconsistencies in the generated images remain uncorrected, and in some local regions the errors become even more severe—falling short of the requirements for consistency-aware correcting.
Another way is to use multi-image editing models, like Qwen-Image~\cite{qwenimage}, to repair local details through text instructions.
As shown in ~\figref{teaser}(a), the model fails to accurately locate and correct the target details.
These demonstrate that, even with a reference image, enhancing the consistency of existing customized generation models remains a challenging problem.

We summarize the failure of the aforementioned models into two main issues.
First, there is a lack of high-quality data that is capable of focusing on and improving fine-grained details.

For example, datasets such as Subjects200K~\cite{ominicontrol}, UNO-1M~\cite{uno}, and X2I2~\cite{wu2025omnigen2} mainly focus on maintaining the overall object consistency while neglecting subtle local details.
To address this, we construct reference–target pairs that maintain strong consistency and simultaneously capture the inherent generation issues of current models. 
Specifically, we employ a VLM-based selection strategy to build the reference–target pairs and further introduce Flux-Fill~\cite{flux} to explicitly degrade local regions of the generated images, thus simulating inconsistency in details and completing our dataset construction.

In addition, existing models often fail to attend to, localize, and align fine-grained regions for precise correction.
To tackle this problem, we visualize the attention maps of the reference and input images within the noise regions and observe strong coupling between them. To mitigate this, we introduce an attention alignment loss, which explicitly supervises the attention maps using region masks, enabling effective decoupling and automatic localization of to be corrected regions.
Furthermore, we analyze the intrinsic properties of our chosen base model.
To improve generalization and reference-image understanding, we proposed a detail encoder, which explicitly embeds features for both reference and input images, fully leveraging the strengths of the pre-trained editing model and further improving the consistency between generated outputs and inputs.

To improve usability, we further design an agent-based processing chain that automates the entire workflow, including consistency evaluation, discrepancy localization, reference retrieval, and image refinement.
This enables multi-round human–agent interaction and a fully automated one-click refinement process, greatly enhancing the model’s ability to handle complex real-world scenarios.

As shown in Fig.~\ref{teaser}, our method significantly improves the consistency of generated images.
The refined results preserve the physical priors of relative spatial relationships between objects while maintaining environmental and lighting consistency.
Meanwhile, local regions such as the wheel hub in Fig.~\ref{teaser}(a) demonstrate a remarkable improvement in structural fidelity and fine-grained alignment.

Our contributions can be summarized as follows. 
\begin{itemize}
\item To further enhance the consistency of existing generative models, we construct a reference-degraded-target dataset and establish a benchmark for complex object customization, providing valuable resources for high-consistency generation and editing research.
\item To address the common issue of detail inconsistency and inaccuracy for existing customized generation models, we introduce an attention alignment loss for region-aware reference perception and a detail encoder for precise task understanding.
\item We design an agent system that seamlessly integrates the critic model, enabling automated and intuitive multi-round editing while ensuring robust performance.
\end{itemize}

\label{sec:dataset}
\begin{figure*}
    \centering
    \includegraphics[width=\linewidth]{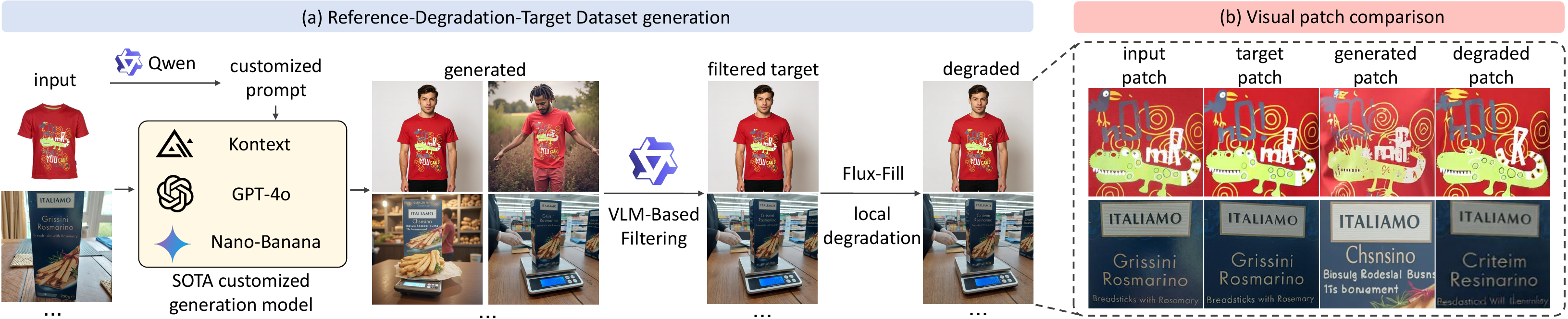}
    \caption{\textbf{Data curation pipeline.} (a) illustrates the complete pipeline of our approach, which involves generating customized images using existing state-of-the-art models, applying VLM-based filtering, and performing degradation. (b) shows local regions from our dataset, where the target patch aligns well with the input patch, and the degraded patch effectively simulates fine-grained inconsistencies in text and logo areas commonly seen between the input patch and the generated patch.
}
\label{fig:data}
\end{figure*}
    
\section{Related Work}
\label{sec:relatedwork}

\myPara{Conditional image generation.} Conditional image generation aims to improve controllability in diffusion models by introducing external conditions. Existing condition-guided diffusion models can be broadly categorized into two types of tasks: \emph{condition-aligned generation} and \emph{condition-referenced generation}.
Condition-aligned generation focuses on abstract structural control, such as pose or shape alignment \cite{zhang2023adding, ma2024followpose, ma2025followyourmotion, ma2024followyouremoji}. In this setting, conditional inputs are strictly aligned with the output at the pixel level, achieving fine-grained spatial control.
In contrast, condition-referenced generation emphasizes high-level semantic and object-specific control rather than pixel-level correspondence. Representative works, such as IP-Adapter~\cite{ye2023ip}, Textual inversion~\cite{gal2022image}, and others, introduce adapters or encoders~\cite{kumari2023multi,wang2024ms,zhang2024ssr,ruiz2023dreambooth,li2024photomaker,gong2025relationadapter} to inject subject-related conditional information into the diffusion process, enabling generation that preserves reference subject consistency.

Unlike previous approaches that inject conditional signals directly at the feature level, DiT-based methods~\cite{le2025one,wang2025unicombine,xia2025dreamomni,xiao2025omnigen,wu2025omnigen2,zhang2025easycontrol,song2025omniconsistency, song2025layertracer, song2025makeanything, wan2024grid, wang2025diffdecompose} unify semantic and spatial conditions into token sequences and integrate them with text tokens through multimodal attention or token concatenation mechanisms.
To enhance subject consistency, DreamO~\cite{dreamo} designs a router mechanism to focus attention on target subjects, and XVerse~\cite{xverse} adopts text-stream modulation to transform reference images into token-specific offsets. MOSAIC~\cite{mosaic} further introduces fine-grained regional constraints via explicit semantic point correspondences.
Despite these advances, inconsistency in fine-grained details remains a significant limitation. To address these issues, we propose a unified post-editing correction framework that substantially enhances the overall consistency of generated images.

\myPara{LLM for Generation.}
Large Language Models (LLMs), such as ChatGPT~\cite{gpt} and Llama~\cite{touvron2023llama,touvron2023llama}, have demonstrated outstanding capabilities in the field of natural language processing. Meanwhile, Multimodal Large Language Models (MLLMs), including LLaVA~\cite{liu2023visual}, Claude, and GPT-4, further integrate visual understanding, significantly enhancing the ability to process multimodal data.
LLMs have now become a core component of vision-language tasks, where their powerful language understanding and reasoning abilities have driven advances in cross-modal semantic alignment and visual reasoning, as demonstrated in works such as VisProg~\cite{gupta2023visual} and ViperGPT~\cite{suris2023vipergpt}.
At the same time, LLMs are widely adopted within agent frameworks~\cite{yang2023auto,shen2023hugginggpt,liu2023interngpt,chen2025re,yuan2025enhancing}, where they learn to invoke external tools to accomplish complex tasks such as visual interaction, speech processing~\cite{gupta2023visual}, software development~\cite{qian2023chatdev}, and game operation~\cite{meta2022human}.
With the deepening of multimodal training, the interconnections among generative agents have become increasingly integrated. LayoutGPT~\cite{feng2023layoutgpt} can generate spatial layouts from textual prompts, GenArtist~\cite{wang2024genartist} and LLM Blueprints employ a ``generate-then-edit'' paradigm to optimize outputs, and LayerCraft~\cite{zhang2025layercraft} introduces an integrated multi-agent framework that unifies layout planning and object integration within a single system.

\section{Data Curation}

% ----------------------

Under our problem setup, the input consists of the customized generated image to be repaired and a reference image. Our goal is to enhance the fine-grained consistency of the generated image.
However, given that existing models often suffer from detail misalignment issues, identifying more consistent input–output pairs to achieve a substantial performance improvement over current models remains a challenging problem.

We decompose the problem into two main components:
constructing consistent reference–target pairs and generating images that reflect the inherent issues present in existing generative models.
First, we employ a VLM-based filtering strategy to leverage the complementary strengths of both inputs, thereby constructing the most ideal ground-truth supervision.
Then, we simulate common artifacts observed in current generative models, such as text rendering errors and logo mismatches, through controlled degradation.

As shown in ~\figref{fig:data}(a), we begin by collecting a set of high-quality images and generating diverse prompts that include various objects, scenes, human–object interactions, and lighting conditions.
Then, by using state-of-the-art text-to-image models, like Flux Kontext~\cite{kontext}, GPT-4o~\cite{gpt}, and Nano-Banana~\cite{nanobanana}, we produce scene-rich images.
These images are evaluated by Qwen-VL~\cite{qwenvl}, which scores them in terms of visual quality and clarity. We retain only the top-performing samples.
Next, Qwen is used again to annotate each image, followed by Grounding SAM~\cite{ren2024grounded} for object detection and segmentation. To mitigate potential mask errors from SAM, we further use Qwen to recheck and filter the outputs, ensuring consistency while preserving segmentation accuracy. This process yields a set of high-quality reference–target pairs.
To simulate real-world degradation, we apply the Flux-Fill~\cite{flux} model to actively corrupt the filtered high-quality images.
Specifically, we randomly select subsets of the masked regions as inputs to Flux-Fill, prompting it to inpaint regions or alter elements, such as text or logos.
The resulting degraded samples are then screened by Qwen to remove those with severe visual errors.

Through this data construction pipeline, we effectively address the two challenges described above and acquire 10k high-quality reference-degraded-target triplets.
As illustrated in ~\figref{fig:data}(b), by comparing the input image patch and the target patch, one can observe that the filtered targets exhibit a high degree of consistency with the original input.
Meanwhile, when comparing the model’s generated patch and the degraded patch, it can be seen that the degraded patches effectively reproduce common artifacts found in current generative models, including logo misalignment and text generation errors, which provide realistic supervision signals and a solid basis for subsequent model training.

\section{Methodology}

\begin{figure*}[ht]
    \centering   
    \includegraphics[width=\linewidth]{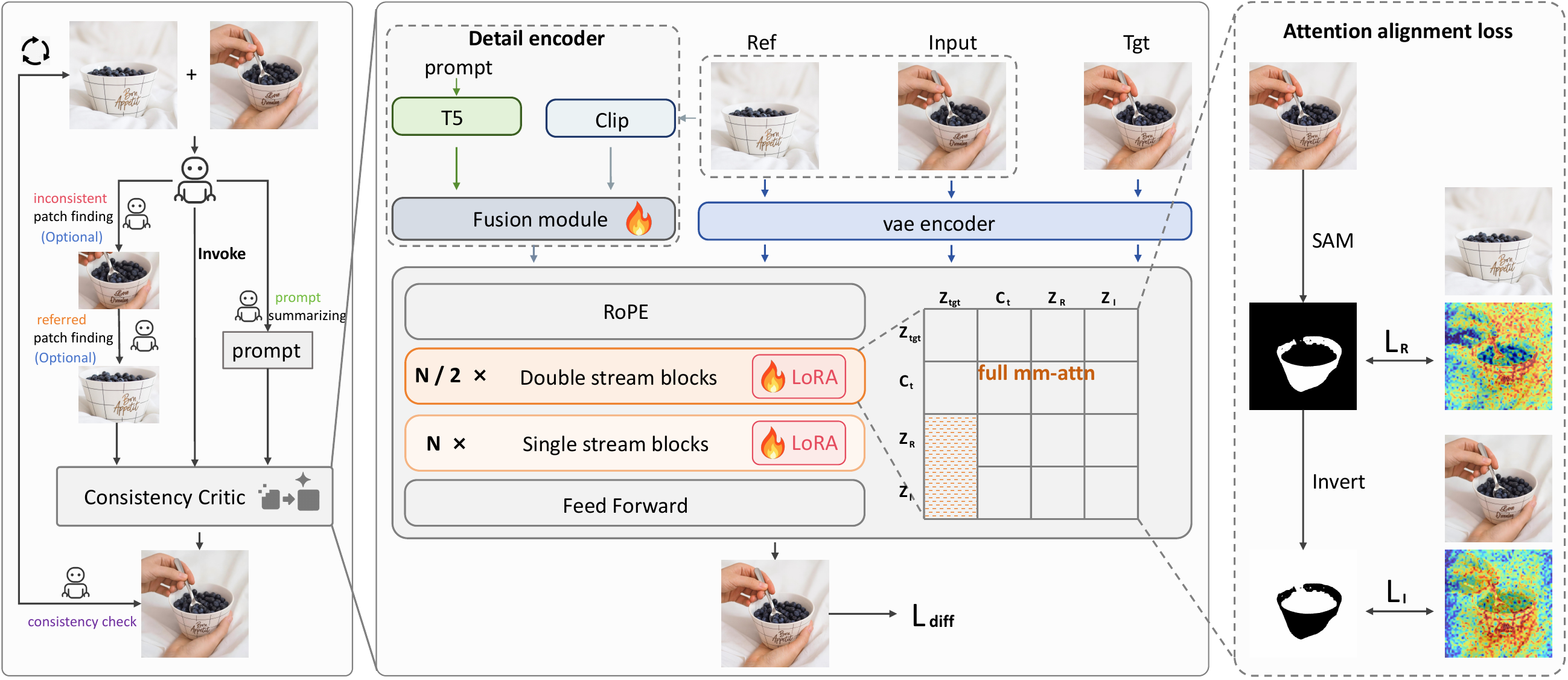}
    \caption{\textbf{Overview of the proposed \ourMthd{}}. We propose \ourMthd{}, which employs a Detail Encoder and an Attention Alignment Loss to enable the model to localize regions requiring restoration, thereby achieving high-quality and consistent image correcting. Furthermore, we develop a fully automated agent framework that supports both local patch restoration and multi-round correcting processes.} 
    \label{fig:net}
    \vspace{-5pt}
\end{figure*}
Our overall pipeline is illustrated in~\figref{fig:net}. As shown, the reference image and input image are fed into the detail encoder together with the prompt to obtain text tokens. These tokens are then concatenated with the image tokens encoded by the VAE and subsequently fed into the DiT for denoising.
During training, we introduce an Attention Alignment Loss to ensure both the disentanglement and alignment of attentions between the condition inputs and the noisy target.
During inference, we design an agent chain that, given a reference image and an input image to be corrected, automatically identifies the required patches and organizes the prompt.
Both training and inference follow the same prompt format: ``Use the {object} in IMG1 as a reference to be corrected, replace, or enhance the {object} in IMG2.'' Here, ``IMG1'' and ``IMG2'' are trigger tokens corresponding to the reference image and the input image, respectively, while \{object\} denotes a simplified description of the target object identified from the reference image.

\subsection{Preliminaries}
\label{sec:Preliminaries}

The DiT architecture employs a multi-modal attention mechanism by concatenating text tokens $c_T \in \mathbb{R}^{M \times d}$ and noisy image tokens $z_{tgt} \in \mathbb{R}^{N \times d}$ as input,
where $d$ denotes the embedding dimension, and $N$ and $M$ are the numbers of image and text tokens, respectively. 
It projects the hidden states of these tokens into query $Q$, key $K$, and value $V$ representations and perform attention:
\begin{equation}
    \text{Attention}(Z) = \text{softmax}\left(\frac{QK^\top}{\sqrt{d}}\right)V,
\end{equation}
where $Z = [z_{tgt}, c_T]$ denotes the concatenation of the text tokens and the noisy image tokens. 
Recent models, such as the Flux Kontext image editing model~\cite{kontext}, extend this formulation to $Z' = [z_{tgt}, c_T, z_c]$, which introduces an additional control token $z_c$ to incorporate image-level conditioning. 
Here, $z_c$ represents an image token corresponding to the image to be edited. We adopt Flux Kontext~\cite{kontext} as our base model to leverage its image control capability for fine-grained consistency correcting.

\subsection{Consistency Critic}
\label{sec:Refiner}

\begin{figure}[ht]
    \centering
    \includegraphics[width=\linewidth]{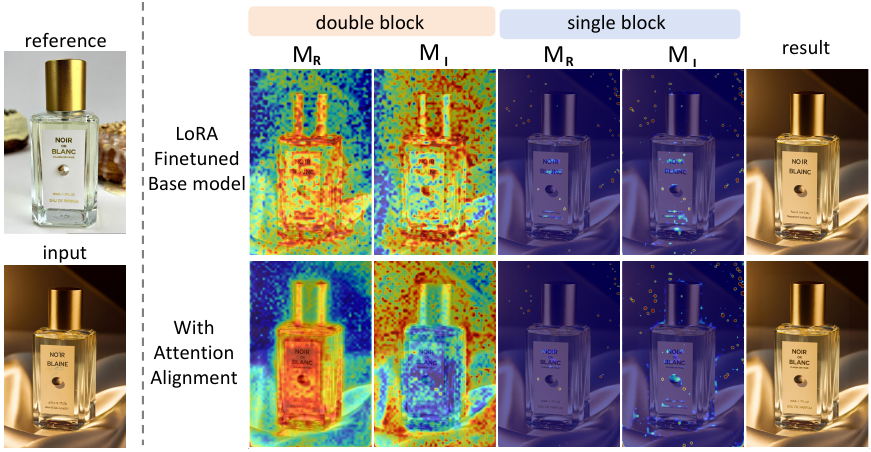}
    \caption{\textbf{Attention visualization.} We separately extract the noise attention maps with respect to the reference image and the input image to be corrected, denoted as $M_R$ and $M_I$, respectively. The first row shows the results of the LoRA-finetuned base model, while the second row presents the results after applying the attention alignment loss. The first two columns correspond to the attention map of the double stream layer, and the last two columns correspond to the single stream layer. It can be observed that the attention alignment loss effectively promotes attention disentanglement.}
    \label{fig:attnmap}
    \vspace{-10pt}
\end{figure}

\myPara{Attention alignment.}
Given the dataset obtained from~\secref{sec:dataset}, a straightforward idea is to fine-tune the base model using LoRA~\cite{lora}. As shown in the first row of~\figref{fig:attnmap}, after LoRA fine-tuning, the model is able to understand our intention and perform local correcting. 
However, for small text or fine-grained regions, the generated images still retain former elements from the input image, indicating that the overall consistency remains insufficient.

As observed in~\cite{he2024uniportrait,he2025anystory,dreamo}, the image-to-image attention mechanism is capable of capturing spatial patterns during generation. Motivated by this, we further visualize the attention maps between the condition inputs and the generation target, as shown in~\figref{fig:attnmap}. 
Specifically, the attention map is formulated as
\begin{equation}
M = \frac{Q_{c_i}K_{tgt}^{T}}{\sqrt{d}}, \quad i \in \{R, I\},
\end{equation}
where $R$ and $I$ denote the reference and input images, respectively, 
$Q_{c_i}$ represents the condition tokens projected into the query space, 
and $K_{tgt}$ denotes the noisy tokens projected into the key space.

By visualizing the influence of tokens from the reference and input branches on the noise latent, we observe that in our fine-tuned model, the attentions from these two branches are strongly coupled within the noise region. As a result, the input and reference tokens convey conflicting cues for local generation, preventing the model from accurately identifying the reference details necessary for precise correction. Consequently, the model either neglects modifications in these regions or performs inaccurate edits.
To achieve the goal of consistency enhancement, we expect the input tokens to provide guidance on object-level details,
while the reference tokens focus on background, lighting, and other global aspects. 
To realize this effect, we introduce an Attention Alignment Loss that enforces alignment at the attention level, explicitly guiding the model to query and reference the target regions appropriately.

Specifically, we first define a binary object mask $B$ as:
\begin{equation}
B(p) =
\begin{cases}
1, & p \in {background}, \\
0, & p \in {subject},
\end{cases}
\end{equation}
Subsequently, we incorporate an MSE-based alignment term into the DiT framework to optimize the attention distributions of both the reference and input branches with respect to the noise regions, defined as:
\begin{equation}
L_{G} = \frac{1}{n_l} 
\sum_{j=0}^{n_l-1} \left\| B \odot  N\!\left(M^j_G\right) \right\|_2^2,
\end{equation}
\begin{equation}
L_{R} = \frac{1}{n_l} 
\sum_{j=0}^{n_l-1} \left\| \overline{B} \odot  N\!\left(M^j_R\right) \right\|_2^2,
\end{equation}
where $\overline{B}$ denotes the complement of $B$, $\odot$ represents element-wise multiplication, $j$ is the layer index, $n_l$ is the total number of layers, and $N(\cdot)$ denotes a min-max normalization operator, which normalizes the input values to the range between zero and one.
Finally, the overall training objective is formulated as:
\begin{equation}
    L = L_{diff}+L_R+L_G.
\end{equation}
Here, $L_{diff}$ represents the rectified flow-matching loss originally adopted by the base model.

This design encourages the reference patches to focus primarily on the object regions rather than the background, while preventing the input patches from exerting excessive influence on the regions that require correction.
Moreover, as observed in~\figref{fig:attnmap}, in the double stream layers, the attentions from both the reference and input branches are concentrated and exhibit strong responses to the noise regions, while in the single block, the attention from the condition branch is more dispersed, lacking distinct regional features.
Therefore, we apply the Attention Alignment Loss only at the double stream blocks of the model, and $n_l$ is set to the number of double-stream block layers only.

\begin{figure}
    \centering
    \setlength{\abovecaptionskip}{2pt}
    \includegraphics[width=1\linewidth]{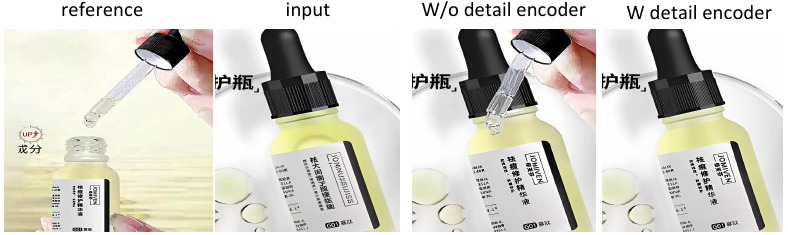}
    \caption{\textbf{Effect of the detail encoder.} 
    We find that when the input image exhibits structural differences from the reference image, the model fails to correctly identify the intended reference object, leading to inconsistent generation results.}
    \label{fig:detail}
    \vspace{-10pt}
\end{figure}

\begin{figure*}[h]
    \centering
    \includegraphics[width=\linewidth]{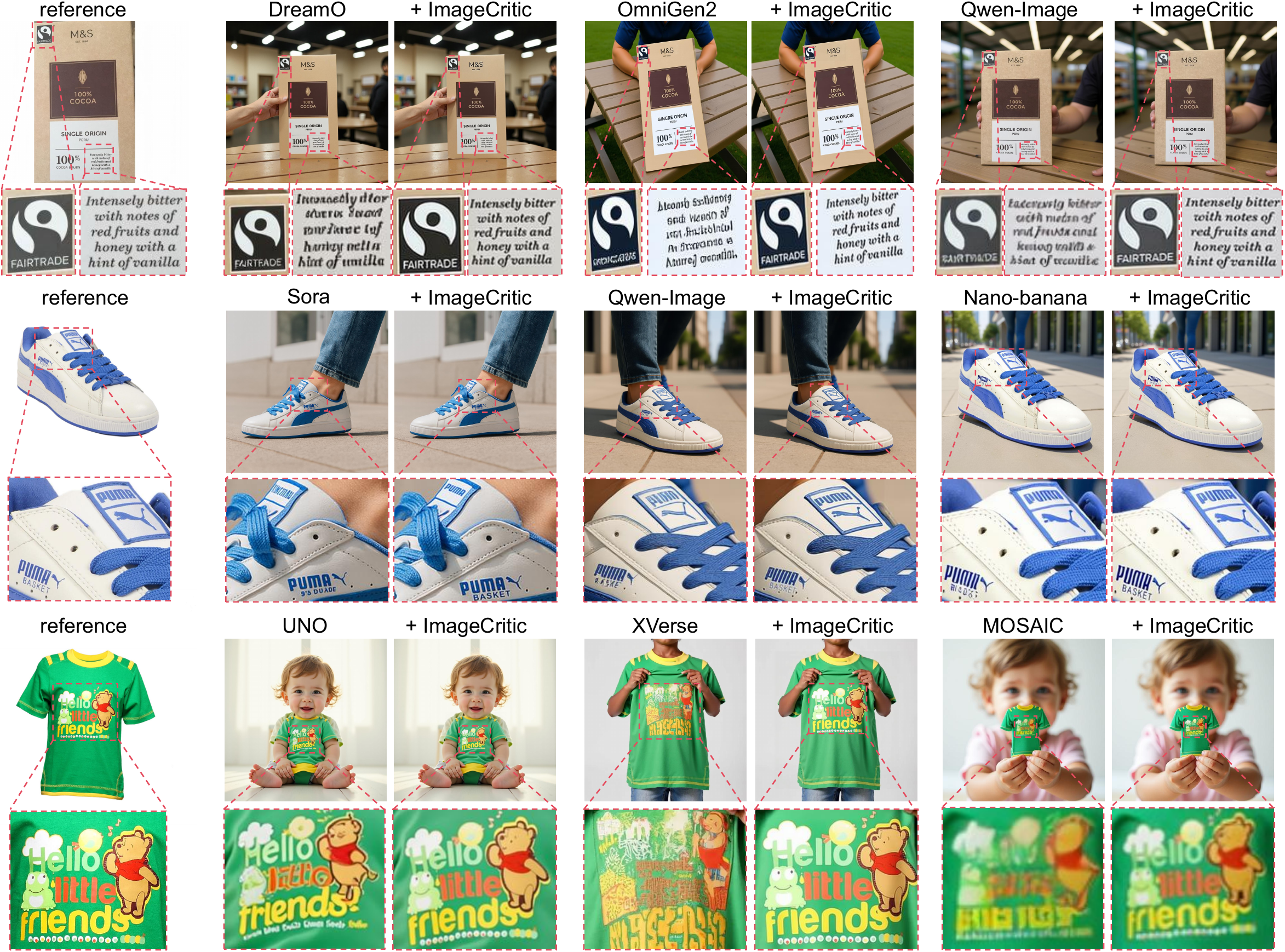}
    \caption{\textbf{Visual Results.} We present the results obtained by applying our method to the generated outputs of existing state-of-the-art customized generation models, including both open-source and closed-source variants. As shown, our method substantially enhances the consistency of the generated images.}
    \label{fig:compare}
    % \vspace{-8pt}
\end{figure*}

\myPara{Detail encoder.}
During training, we provide the model with two image inputs simultaneously and set the triggers ``IMG1'' and ``IMG2" to represent each image, enabling precise prompt description.
However, for the T5 text encoder, the latent representation derived from the same trigger word remains identical across different inputs, which causes the model to struggle in associating different image inputs with their respective triggers, leading to incorrect outputs.
As shown in~\figref{fig:detail}, when the input image slightly differs in shape from the reference image, the model struggles to accurately identify the corresponding part of both condition images within the prompt, leading to erroneous generation.

Inspired by works such as PhotoMaker~\cite{li2024photomaker} and Dreamo~\cite{dreamo}, we design a Detail Encoder that couples the latent representation of the trigger word with that of the image, enhancing the model’s understanding of the reference image.
This coupling mechanism allows the model to link textual triggers and visual content more effectively, thereby improving reference consistency in the generated results.

Specifically, let the prompt, after being processed by the T5 encoder, produce token representations 
$P \in \mathbb{R}^{M \times d_t},$
and let the flattened hidden representation of the image obtained from CLIP~\cite{radford2021learning} be 
$C_i \in \mathbb{R}^{1 \times d_c}.$
From these tokens, we extract the hidden states 
$P_R, P_I \in \mathbb{R}^{1 \times d_t},$
corresponding to the trigger words ``IMG1'' and ``IMG2'', which refer to the reference image and the input image to be corrected, respectively. 
Here, $d_t$ and $d_c$ denote the hidden dimensionalities of the T5 and CLIP models, respectively.
We then construct a fused hidden representation:
\begin{equation}
    P_i' = [P_i; C_i] \in \mathbb{R}^{1 \times (d_t + d_c)}, \quad i \in \{R, I\}.
\end{equation}
% where $i$ corresponds to the reference (\(R\)) and input (\(I\)) images. 
Subsequently, $P_i'$ is passed through a two-layer MLP and projected back to the original dimension, yielding $\tilde{P}_i$, which is then used to update the corresponding hidden states of the trigger words $P_R$ and $P_I$ in the prompt representation, respectively.

\subsection{Agent Chain}
\label{sec:agentchain}
In our experiments across various generative models, we observe that they tend to compress the reference image, resulting in local regions where fine textual details become too small or even fail to render properly.
However, in real-world scenarios, users often provide high-resolution reference images that are not fully utilized by such models.
Moreover, existing models sometimes produce results in which the reference image occupies only a small portion of the generated output, which can negatively affect the correcting quality.
To better preserve fine details, fully leverage high-resolution guidance, and enable more intuitive interaction, we develop an agent chain framework.
Our system integrates multiple specialized agents through carefully designed prompts, with each agent responsible for specific subtasks—such as assessing content consistency, identifying inconsistent regions, selecting relevant reference patches, summarizing prompts, and performing final image correction (handled by the critic model).
Specifically, we employ the Qwen-Agent~\cite{qwenvl} as a coordinator, which supervises the overall workflow, facilitates smooth communication between users and agents, and orchestrates cooperation among all components. 
While the system can operate autonomously, users can also interactively adjust input patches.
For instance, if a user wishes to regenerate small textual details within a product region, the coordinator delegates this request to the corresponding agent for reprocessing, ensuring that the input patches align precisely with the user’s intent.

\begin{figure}[t]
    \centering
    \includegraphics[width=\linewidth]{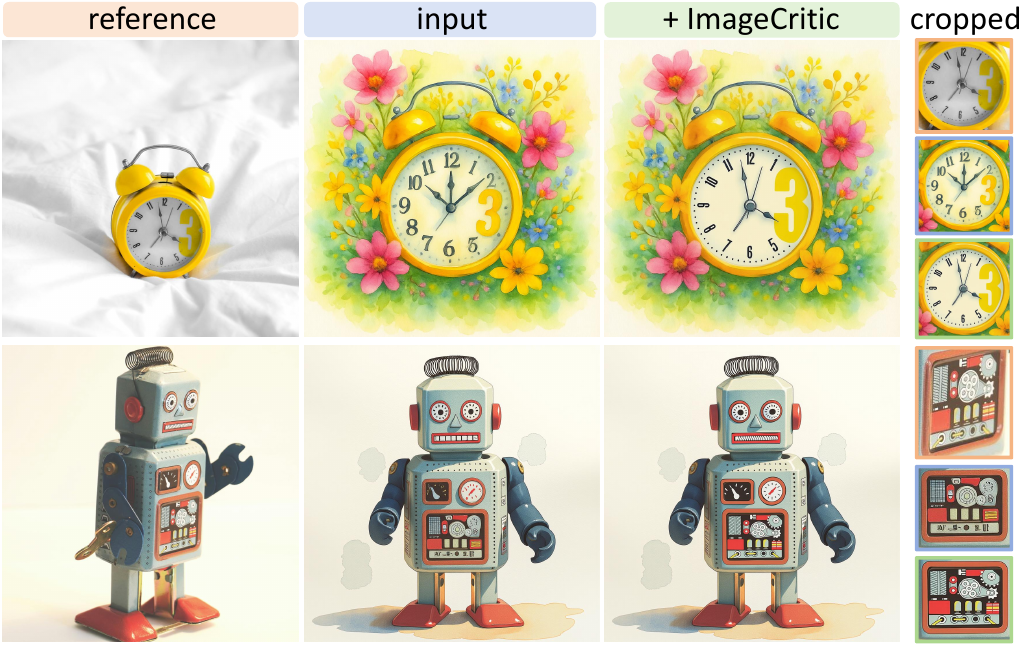}
    % \vspace{-14pt}
    \caption{\textbf{Robust Generalization.} As shown in the clock’s numerals and hands and the robot’s panels, our proposed \ourMthd{} effectively preserves style and illumination consistency rather than relying on naive copy-paste patterns, achieving detailed and coherent corrections across varying viewpoints, cross-category objects, and diverse stylization settings.}
    \label{fig:Robust}
    % \vspace{-4pt}
\end{figure}

\section{Experiments}

\subsection{Implementation details.}
We adopt Flux.1-Kontext-dev~\cite{kontext} as our base model, upon which we conduct LoRA fine-tuning with a learning rate of $1 \times10^{-4}$ and a rank of 128, training on 2 GPUs with a batch size of 4 per GPU (total batch size = 8) for a total of 20,000 steps.
After training, we enhance a subset of the generation targets with low consistency in the dataset using our proposed \ourMthd{} to further improve overall generation consistency. Details are provided in the Supplementary Material. The model was then retrained following the same settings described above.

\subsection{Qualitative comparisons.}

We employ both open-source models — XVerse~\cite{xverse}, Dreamo~\cite{dreamo}, MOSAIC~\cite{mosaic}, OmniGen2~\cite{wu2025omnigen2}, UNO~\cite{uno}, and Qwen-Image~\cite{qwenimage} — as well as the closed-source models NanoBanana~\cite{nanobanana} and GPT-Image~\cite{gpt}, to evaluate single subject customization performance.
Subsequently, we apply our proposed method to enhance the consistency of the generated results.
As illustrated in~\figref{fig:compare}, existing approaches can generally preserve the primary subject; however, they often struggle to handle fine-grained local details, which limits their overall applicability.
In contrast, our method achieves a significant improvement in consistency while maintaining the original lighting, texture, and background fidelity of the generated images.
In addition to these advantages, our approach further demonstrates strong robustness and generalization across diverse scenarios.
As shown in~\figref{fig:Robust}, our method achieves highly consistent correcting across varying viewpoints, object categories, and diverse stylization settings, while faithfully preserving the background, illumination, and stylistic characteristics of the input images.

\subsection{Quantitative comparisons.}
\label{quantitative}

\myPara{Benchmark.}
We first evaluate our model on Dreambench++~\cite{dreambench}.
Subsequently, to further examine the model’s performance on more challenging cases involving text, logos, and other complex visual details, we introduce a new benchmark, \ourBench{}, which specifically focuses on fine-grained detail preservation.
Our benchmark consists of 300 images, including 200 intricate multilingual product images and 100 images of apparel and accessories.
% 
% For each image, we use the Qwen-VL~\cite{qwenvl} model to generate editing instructions that best match the input image’s intended scene.
For each image, we utilize the Qwen-VL~\cite{qwenvl} model to generate editing instructions that most accurately align with the intended scene of the input image.
Subsequently, we apply our method to these results and evaluate the quality of the final generated images.

\myPara{Evaluation Metrics.}
\begin{table}[t]
\centering
\scriptsize
\setlength{\tabcolsep}{5pt}
\caption{\textbf{Quantitative comparison on CriticBench.} Evaluation of consistency correcting across different generation models on our proposed CriticBench.}
\vspace{-6pt}
\label{tab:bench1}
\begin{tabular}{l|ccc}
\toprule
 \textbf{Method} & \textbf{CLIP-I~$\uparrow$} & \textbf{DINO~$\uparrow$} & \textbf{DreamSim~$\downarrow$} \\
\midrule
Sora~\cite{gpt} & \incarr{78.7}{79.6} {\small{\textcolor{myblue}{\scriptsize {+0.9}}}} & \incarr{68.4}{69.2} {\small{\textcolor{myblue}{\scriptsize {+0.8}}}} & \incarr{29.1}{28.7} {\small{\textcolor{myblue}{\scriptsize {-0.4}}}} \\
Nano-Banana~\cite{nanobanana} & \incarr{79.2}{79.8} {\small{\textcolor{myblue}{\scriptsize {+0.6}}}} & \incarr{66.5}{66.9} {\small{\textcolor{myblue}{\scriptsize {+0.4}}}} & \incarr{32.0}{31.8} {\small{\textcolor{myblue}{\scriptsize {-0.2}}}} \\
\midrule
Xverse~\cite{xverse} & \incarr{76.5}{79.9} {\small{\textcolor{myblue}{\scriptsize {+3.4}}}} & \incarr{68.8}{71.9} {\small{\textcolor{myblue}{\scriptsize {+3.1}}}} & \incarr{34.3}{31.4} {\small{\textcolor{myblue}{\scriptsize {-2.9}}}} \\
DreamO~\cite{dreamo} & \incarr{77.8}{78.1} {\small{\textcolor{myblue}{\scriptsize {+0.3}}}} & \incarr{67.7}{68.2} {\small{\textcolor{myblue}{\scriptsize {+0.5}}}} & \incarr{29.6}{29.2} {\small{\textcolor{myblue}{\scriptsize {-0.4}}}} \\  
MOSAIC~\cite{mosaic} & \incarr{74.6}{77.1} {\small{\textcolor{myblue}{\scriptsize {+2.5}}}} & \incarr{62.6}{65.0} {\small{\textcolor{myblue}{\scriptsize {+2.4}}}} & \incarr{35.2}{31.4} {\small{\textcolor{myblue}{\scriptsize {-3.8}}}} \\
OmniGen2~\cite{wu2025omnigen2} & \incarr{78.8}{79.3} {\small{\textcolor{myblue}{\scriptsize {+0.5}}}} & \incarr{70.0}{70.8} {\small{\textcolor{myblue}{\scriptsize {+0.8}}}} & \incarr{27.7}{27.0} {\small{\textcolor{myblue}{\scriptsize {-0.7}}}} \\
UNO~\cite{uno} & \incarr{77.6}{78.9} {\small{\textcolor{myblue}{\scriptsize {+1.3}}}} & \incarr{68.4}{69.3} {\small{\textcolor{myblue}{\scriptsize {+0.9}}}} & \incarr{33.6}{32.1} {\small{\textcolor{myblue}{\scriptsize {-1.5}}}} \\
Qwen-Image~\cite{qwenimage} & \incarr{77.9}{78.2} {\small{\textcolor{myblue}{\scriptsize {+0.3}}}} & \incarr{69.2}{69.4} {\small{\textcolor{myblue}{\scriptsize {+0.2}}}} & \incarr{30.3}{30.1} {\small{\textcolor{myblue}{\scriptsize {-0.2}}}} \\
\bottomrule
\end{tabular}
\vspace{-3pt}

\end{table}

\begin{table}[t]
\centering
\scriptsize
% \vspace{4pt}
\setlength{\tabcolsep}{5pt}
\caption{\textbf{Quantitative comparison on DreamBench++.} Evaluation of consistency correcting across different generation models on DreamBench++.}
\vspace{-6pt}
\label{tab:bench2}

\begin{tabular}{l|ccc}
\toprule
\textbf{Method} & \textbf{CLIP-I~$\uparrow$} & \textbf{DINO~$\uparrow$} & \textbf{DreamSim~$\downarrow$} \\
\midrule
Xverse~\cite{xverse} & \incarr{81.5}{82.3} {\small{\textcolor{myblue}{\scriptsize {+0.8}}}} & \incarr{64.5}{66.3} {\small{\textcolor{myblue}{\scriptsize {+1.8}}}} & \incarr{36.9}{34.9} {\small{\textcolor{myblue}{\scriptsize {-2.0}}}} \\
Dreamo~\cite{dreamo} & \incarr{83.3}{83.4} {\small{\textcolor{myblue}{\scriptsize {+0.1}}}} & \incarr{72.5}{72.7} {\small{\textcolor{myblue}{\scriptsize {+0.2}}}} & \incarr{27.2}{27.1} {\small{\textcolor{myblue}{\scriptsize {-0.1}}}} \\  
MOSAIC~\cite{mosaic} & \incarr{77.5}{78.3} {\small{\textcolor{myblue}{\scriptsize {+0.8}}}} & \incarr{57.6}{59.2} {\small{\textcolor{myblue}{\scriptsize {+1.6}}}} & \incarr{42.7}{41.1} {\small{\textcolor{myblue}{\scriptsize {-1.6}}}} \\
OmniGen2~\cite{wu2025omnigen2} & \incarr{78.5}{79.2} {\small{\textcolor{myblue}{\scriptsize {+0.7}}}} & \incarr{62.0}{64.5} {\small{\textcolor{myblue}{\scriptsize {+2.5}}}} & \incarr{38.6}{36.3} {\small{\textcolor{myblue}{\scriptsize {-2.3}}}} \\
UNO~\cite{uno} & \incarr{78.7}{79.0} {\small{\textcolor{myblue}{\scriptsize {+0.3}}}} & \incarr{62.7}{63.2} {\small{\textcolor{myblue}{\scriptsize {+0.5}}}} & \incarr{37.7}{37.1} {\small{\textcolor{myblue}{\scriptsize {-0.6}}}} \\
Qwen-Image~\cite{qwenimage} & \incarr{78.8}{79.7} {\small{\textcolor{myblue}{\scriptsize {+0.9}}}} & \incarr{61.8}{64.9} {\small{\textcolor{myblue}{\scriptsize {+3.1}}}} & \incarr{36.9}{34.5} {\small{\textcolor{myblue}{\scriptsize {-2.4}}}} \\
\bottomrule
\end{tabular}

% \vspace{-3pt}

\end{table}

We evaluate our approach on the aforementioned DreamBench++~\cite{dreambench} and \ourBench{} benchmarks, leveraging the methods introduced earlier for comparison.
For each method, we compare each generated image with its corresponding reference image and compute metrics, including CLIP Image Score~\cite{radford2021learning}, DINO Score~\cite{zhang2022dino}, and DreamSim~\cite{fu2023dreamsim}.
Specifically, on our proposed \ourBench{}, for each method, we utilize Qwen-VL~\cite{qwenvl} to perform semantic grounding on the corresponding objects within the generated images to enhance the evaluation of fine-grained detail consistency.
The evaluation results on DreamBench++~\cite{dreambench} and \ourBench{} are presented in~\tabref{tab:bench1} and ~\tabref{tab:bench2}, respectively. 
As shown, our method substantially improves the consistency between the generated images and their corresponding reference images, demonstrating strong capability in recovering details.
\subsection{Ablation analysis}
\begin{table}[t]
\centering
\scriptsize
% \resizebox{0.75\linewidth}{!}{ % Resize the table to fit the text width
\setlength{\tabcolsep}{10pt} % myblueuce the table column space
% \vspace{-6pt}
\caption{\textbf{Ablation study.} Quantitative evaluation of the contribution of each component in our proposed model.}
\vspace{-6pt}
\label{tab:ablation}
\begin{tabular}{cc|ccc}
\toprule
{\textbf{AAL}} & {\textbf{DE}}  & \textbf{CLIP-I} & \textbf{DINO} & \textbf{DreamSim} \\
\midrule
\xmark & \xmark  & 77.9 {\small{\textcolor{myblue}{\scriptsize {+0.3}}}} & 68.1 {\small{\textcolor{myblue}{\scriptsize {+0.4}}}} & 31.3 {\small{\textcolor{myblue}{\scriptsize {-0.2}}}} \\ 
\xmark & \cmark  & 78.3 {\small{\textcolor{myblue}{\scriptsize {+0.7}}}} & 68.4 {\small{\textcolor{myblue}{\scriptsize {+0.7}}}} & 30.6 {\small{\textcolor{myblue}{\scriptsize {-0.9}}}} \\ 
\cmark & \xmark  & 78.3 {\small{\textcolor{myblue}{\scriptsize {+0.7}}}} & 68.6 {\small{\textcolor{myblue}{\scriptsize {+0.9}}}} & 30.6 {\small{\textcolor{myblue}{\scriptsize {-0.9}}}} \\ 
\cmark & \cmark  & 78.9 {\small{\textcolor{myblue}{\scriptsize {+1.3}}}} & 68.9 {\small{\textcolor{myblue}{\scriptsize {+1.2}}}} & 29.8 {\small{\textcolor{myblue}{\scriptsize {-1.7}}}} \\ 
\bottomrule
\end{tabular}
% }

\vspace{-10pt}

\end{table}

We validate the effectiveness of our proposed dataset and conduct ablation studies on two key design components: Attention Alignment Loss (AAL) and Detail Encoder (DE).

\myPara{Qualitative comparisons.} As presented in ~\figref{fig:attnmap} and~\figref{fig:detail}, We compare the generation results of the model before and after adding our proposed module.
As shown in~\figref{fig:attnmap}, the results of the fine-tuned base model validate the effectiveness of our proposed dataset. However, simple fine-tuning still leads to deficiencies in subtle local details. After introducing the AAL, the attentions from the input and reference branches become concentrated on the required object and background regions, achieving attention disentanglement and significantly improving the model’s ability to correct local details.
Meanwhile, as shown in~\figref{fig:detail}, after incorporating the DE, the model’s capability to interpret complex inputs is enhanced, enabling it to accurately locate the target regions.

\myPara{Quantitative evaluation}, Following the setup in~\secref{quantitative}, we perform consistency correcting on the generated results of different methods using \ourBench{}, and report the average score improvements across all methods.
As shown in ~\tabref{tab:ablation}, fine-tuning the base model on our dataset alone already improves the consistency of generated images, verifying the effectiveness of our proposed dataset.
Moreover, introducing either component individually leads to additional performance gains, while combining both yields a further and more substantial improvement.
These results confirm the effectiveness of our proposed modules and their synergistic effect.

\subsection{Discussion}
To evaluate the effectiveness of our proposed agent chain in locating input and reference patches, we manually annotated the target regions requiring correction along with their corresponding reference regions within the proposed benchmark. We then compared these human-defined bounding boxes with those automatically predicted by our agent, and quantified their alignment using the Intersection over Union (IoU):
\begin{equation}
\text{IoU} = \frac{|B_{\text{agent}} \cap B_{\text{human}}|}{|B_{\text{agent}} \cup B_{\text{human}}|}, 
\label{eq:iou}
\end{equation}
where $B_{\text{agent}}$ and $B_{\text{human}}$ represent the bounding boxes predicted by the agent and human annotators, respectively. Additionally, we also compute the mean Average Precision at a 50\% IoU threshold (mAP@50), which assesses how accurately the agent detects and localizes regions whose overlap with the ground truth exceeds 50\%.
As shown in ~\tabref{tab:agent_iou_map}, our agent achieves a mean IoU of \textbf{75.3\%} and a mean mAP@50 of \textbf{88.4\%}, indicating its strong capability in accurately querying and localizing fine-grained details.

\begin{table}[t]
\centering
\scriptsize
\renewcommand{\arraystretch}{1.1}
\setlength{\tabcolsep}{10pt}
\caption{\textbf{Agent chain discussions.} Comparison between our proposed agent chain and human annotations in locating required input and reference patches.}
\vspace{-6pt}

\label{tab:agent_iou_map}
\begin{tabular}{c|cc}
\toprule
\textbf{Metric} & \textbf{Mean IoU (\%)} & \textbf{mAP@50 (\%)} \\
\midrule
\textbf{Score} & 75.3 & 88.4 \\
\bottomrule
\end{tabular}
\vspace{-10pt}
\end{table}

\section{Conclusions}

In this paper, we introduce \ourMthd{}, a novel post-editing correction framework designed to enhance the consistency of customized image generation, and construct a reference-degraded-target triplet dataset tailored to this task, enabling more accurate correcting.
Extensive experimental results demonstrate that \ourMthd{} effectively mitigates inconsistencies in generated images, significantly improving detail consistency across both open-source and closed-source models.

{\small
\bibliographystyle{ieee_fullname}
\bibliography{egbib}
}

% \newpage
\appendix
\clearpage

\begin{center}
     \Large\textbf{Supplementary Material}
     % \\\textbf{\ourMthd{}: Unlocking Training-Free Fusion of Any Subject and Style LoRAs}}
\end{center}

\noindent The supplementary material is structured as follows:
% 我们的数据集中的case以及和其余数据集的对比 
% 我们展示了更多数据集构造中的细节 in ~\secref{degrada}
% 我们对比了更多现有的编辑方法在修复任务中的表现 in ~\secref{addition comparisions}
% 我们展示了our proposed XX 在多语言 多方法下的生成结果   ~\secref{addition comparisions}
% 我们展示了agent chain中采取的llm prompt

\begin{enumerate}
    \item We provide additional details on the construction of our dataset in~\secref{degradation}.
    \item We compared additional existing editing methods on the restoration task in~\secref{addition comparisons}.
    \item We provide representative cases from our dataset and compare them with those from other existing datasets in~\secref{dataset_comparison}.
    \item We present the correction results of our proposed \ourMthd{} across multiple languages on images generated by different models in~\secref{visual}.
    \item We present an example of using the agent chain and the prompts used by each of our agent components in~\secref{agent}.

\end{enumerate}

\section{Data Curation details} \label{degradation}

\begin{figure*}
    \centering
    \includegraphics[width=1\linewidth]{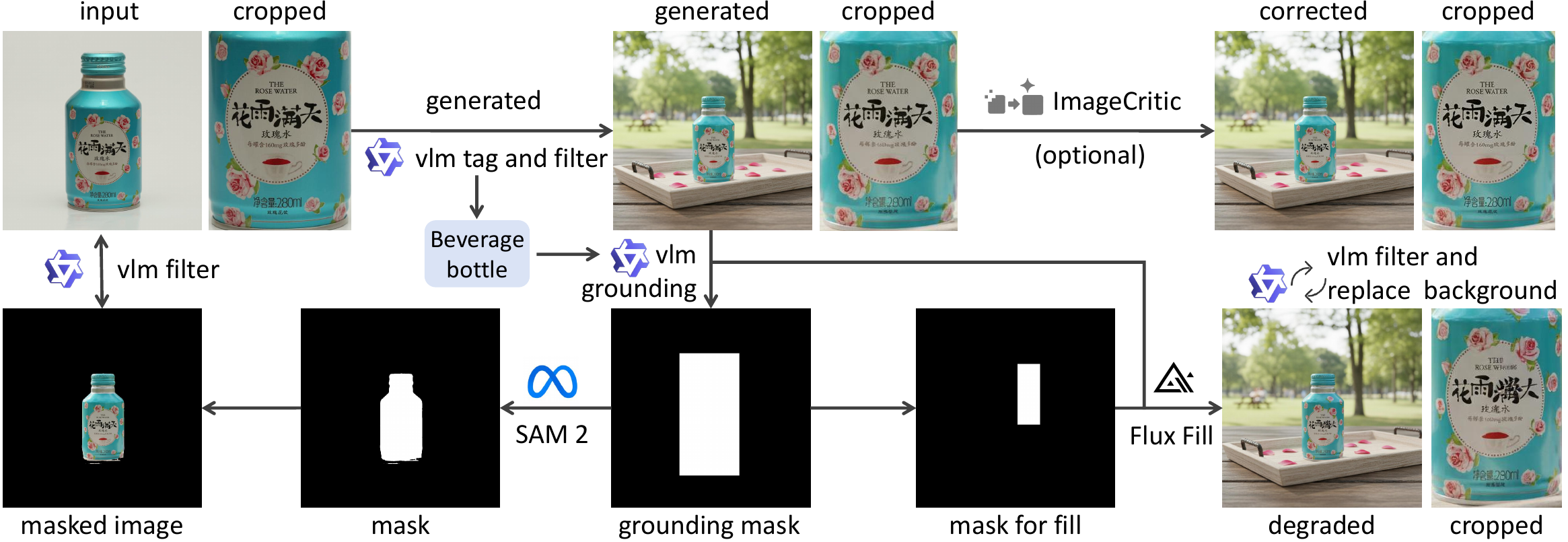}
    % 我们展示了我们在数据构建和退化中使用细节
    \caption{\textbf{Data curation details.} We present a comprehensive account of our dataset construction methodology, encompassing data generation, data annotation, and data augmentation procedures.}

    \label{data}
\end{figure*}

% 在通过爬取下载的方式获取了大量商品数据集后 我们首先使用sota的生成方法构造生成图片，随后使用Vlm进行一致性筛选 删除不一致的pair
% prompt为
% 随后我们使用vlm对生成图片进行tag的方式，以获取了物体的类别 用于训练和grounding 
% 为了保证模型能正确理解复杂场景中的语义 我们使用qwen vlm模型并根据product参考 实现对生成图片中对应物体的grounding
% prompt为“”
% 随后 根据vlm模型生成的 grounding bbox， 我们使用sam对生成的图片进行检测 获取具体的mask，
% 为了进一步确保mask的正确性并进一步提升生成的一致性 我们再一次使用vlm，对比full mask和product中 接着 prompt为“”
% 随后我们根据物体整体区域的 full mask，然后从中随机抽取 20%～70% 的局部区域 生成矩形区域 用于flux fill模型的掩码输入 随后 我们为flux fill设置三种prompt-english words chinese words or logos 或者输入空prompt 以适配不同场景
% 接着 我们再次使用vlm 筛选出flux fill中部分严重退化的图片
% prompt为“”
% 最后 令获取的物体M 退化后的图片D 生成后的图片G 我们最终的退化图片为 M*D + （1-M）*G  用这种方式  排除了fill退化过程中可能在背景区域产生的伪影 
% 从数据层面保证了退化仅在物体的内部发生而不会影响到原先的背景环境
As illustrated in~\figref{data}, our data construction pipeline proceeds sequentially as follows.

\subsection*{Step 1: Reference-target pairs generation}
We begin by collecting a large-scale product dataset through web crawling and downloading and employ state-of-the-art generative models to produce synthetic image variants for each product sample.

\subsection*{Step 2: Generation quality filtering}
We then used Qwen3-vl~\cite{qwenvl} to perform quality filtering with prompt: \textit{\uline{``Determine whether the text details in the input image are both strictly clearly visible and fully readable. If any part of the text is blurred, low-resolution, or difficult to recognize, answer `No'. Otherwise, answer `Yes'. Respond with only `Yes' or `No', followed by a brief reason.''}}.

\subsection*{Step 3: Semantic tags generation}
Next, we employed Qwen~\cite{qwenvl} to assign semantic tags to the generated images in order to extract object categories for training purposes with the prompt \textit{\uline{``Given the image of an object, return only the most general category of the object using exactly 1 to 3 words. Strictly avoid any additional details or descriptions."}}.

\subsection*{Step 4: Image Grounding}
To ensure that the model correctly interprets semantics in complex scenes, we utilized the Qwen~\cite{qwenvl} for grounding. The prompting strategy was as follows: \textit{\uline{``Given image1 as a reference, detect the same object in image2 and output the bounding box coordinates strictly in the format [x1, y1, x2, y2], where x1,y1 are the top-left integers and x2,y2 are the bottom-right integers. You must return only the coordinates in that exact format with no extra text.''}} where image1 and image2 referred to the input image and the generated image, respectively.

\subsection*{Step 5: Consistency Verification}
Based on the bounding boxes predicted by Qwen~\cite{qwenvl}, we adopted SAM~\cite{ravi2024sam2segmentimages} to detect and extract precise object masks. To further guarantee the accuracy of these masks and enhance generative consistency, we invoked the Qwen again to compare the full-object masks against the product references, using the following prompt: \textit{\uline{``Please analyze if the extracted region in Image 1 corresponds to the product in Image 2. Ignore the background and focus only on the main object. 1. Are the objects in both images the same product? If Image 2 contains only a small portion of a local region from Image 1, consider it as No. (Yes/No). 2. Explain based on visual features such as shape, color, texture, and context. If the object in Image 2 contains the mask region, describe the match.''}} where image1 and image2 referred to the input image and the generated image, respectively.

\subsection*{Step 6: Image degradation}
From each verified full-object mask, we sampled 20\%–70\% of the region to obtain a random local rectangular mask used as the inpainting region for the Flux-Fill~\cite{flux} model. For Flux-Fill, we designed three types of prompts---\textit{\uline{``English words''}}, \textit{\uline{``Chinese characters''}}, \textit{\uline{``logos''}}---as well as an empty-prompt setting to accommodate different scenarios.

\subsection*{Step 7: Degradation quality filtering}
After generation, we again used a Qwen~\cite{qwenvl} to filter out severely degraded results produced by Flux-Fill, with the following prompt: \textit{\uline{``Is there is obvious distorted text or mismatched elements in the image. Answer with 'Yes' or 'No', followed by a brief explanation of the reason''}}.

\subsection*{Step 8: Final Image Composition}
Let $M$ denote a binary full-object mask generated by SAM~\cite{ravi2024sam2segmentimages}, $D$ the degraded image, and $G$ the generated image.
We construct the final degraded sample by combining the degraded content inside the object region with 
the generated content outside it:
\begin{equation}
    I_{\text{final}} = M \cdot D + (1 - M) \cdot G .
\end{equation}
This procedure effectively suppresses artifacts that may arise in background areas during the filling process, ensuring that degradation occurs strictly within the object region without affecting the original background environment.

\subsection*{Step 9: Iterative Data Enhancement Strategy}
% 在按照上述方法对我们的模型进行 训练之后 我们的模型已经可以对局部进行修复
% 因此 为了进一步提高生成的一致性和质量 我们对其中部分数据做了增强，具体而言 由于我们在构造数据阶段获取了 对物体的grounding mask，我们将对应区域裁剪下来 和输入图片一起送如我们的方法 进行一致性的修复，并用修复后的图片作为target更新数据集 进一步提升数据质量
After following the training procedure described above, our model is able to perform local region restoration. To further improve the consistency and quality of the generated outputs, we apply additional enhancement to the generated image with Chinese characters. Specifically, since we obtained grounding masks for objects during data construction, we crop the corresponding regions and feed them—together with the original input image—into our method for consistency correction. The corrected images are then used as new targets to update the dataset, thereby further improving overall data quality.

\section{Image Correction Comparison}\label{addition comparisons}
% 我们进一步对比了更多 多模态大语言模型 如 nanobanana gpt-image 1和编辑模型如 Omnigen2 Qwen-imag 对生成图片进行一致性修复任务中的能力
% 特别的 为了  更好让模型理解我们需要的任务 我们设计了如下的prompt 
% 'use the product in the left first reference image as a reference to refine, replace, enhance the product in the right second to-be-refined image, matching their texture, details, color, logo, and texts, while preserving everything else in the right second to-be-refined image untouched'并展开测试
% 如图所示 现有的方法无法在保证背景一致性的情况下 对生成的图片进行一致性correcting，而进过我们的方法 图片在保持背景的情况下 一致性大幅度提升
We further compared additional multimodal large language models, such as Nano-banana~\cite{nanobanana} and GPT-Image 1~\cite{gpt}, as well as editing models like Omnigen2~\cite{wu2025omnigen2} and Qwen-image~\cite{qwenimage}, evaluating their capabilities on the task of consistency correction in generated images.

\begin{figure*}
    \centering
    \includegraphics[width=1\linewidth]{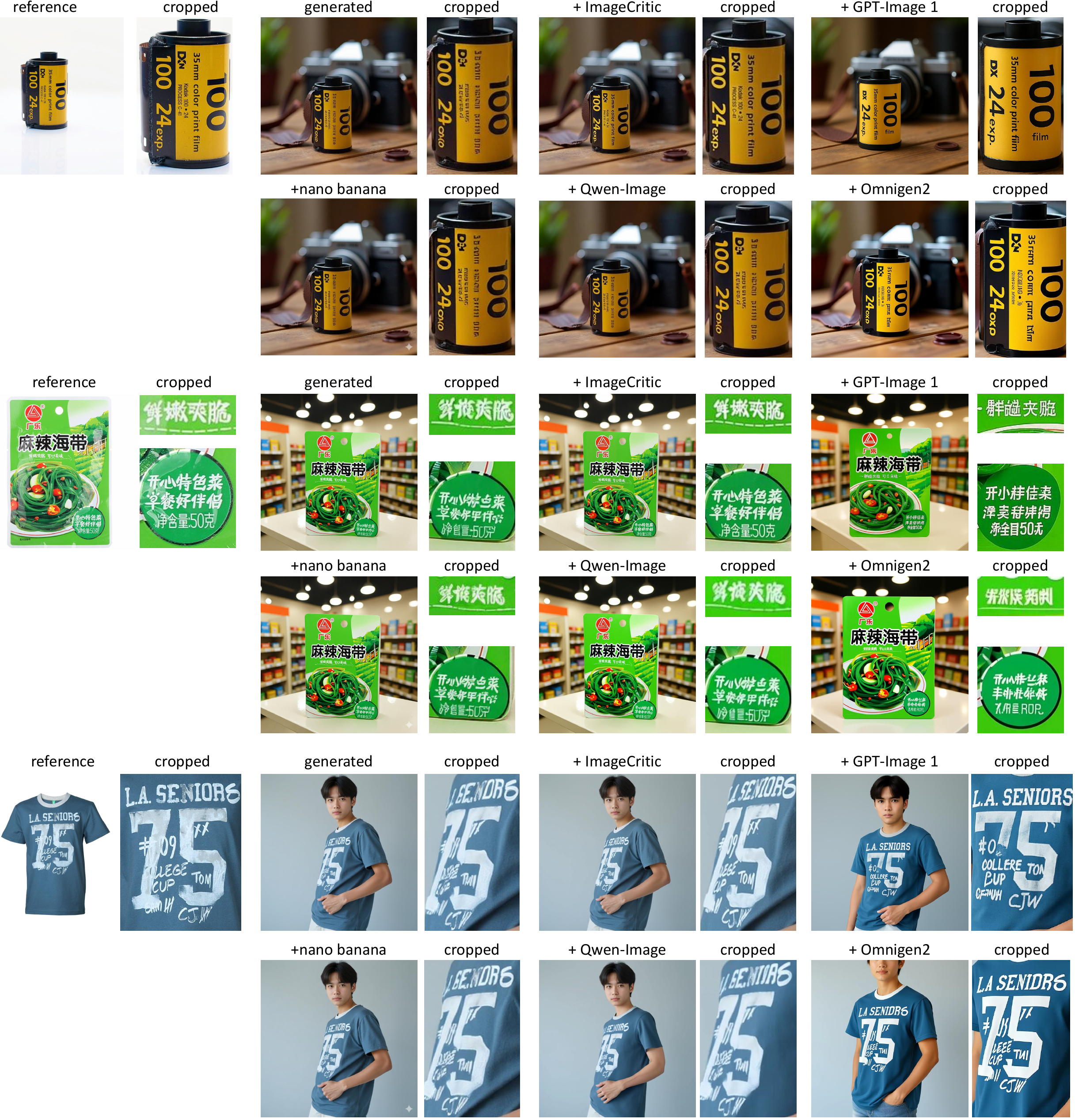}
    % Additional Comparisons. 我们测试了现有的多模态方法和编辑方法 从图中可以看出 现有模型在保持整体一致性或局部修复中存在问题 影响了实际的使用，相比而言 我们的方法在保持生成局部一致性的同时 保证了背景的一致性 
    \caption{\textbf{Comparisons with multimodal and editing model}. We evaluated existing multimodal and editing methods. As shown, current models exhibit issues in preserving global coherence or performing localized corrections, which affects their practical applicability. In contrast, our approach maintains local consistency in the generated content while ensuring overall background coherence.}
    \label{fig:othermethod}
\end{figure*}

To help the models better understand the task, we designed the following prompt:
\textit{\uline{``Use the product in the left first reference image as a reference to refine, replace, and enhance the product in the right second to-be-refined image, matching their texture, details, color, logo, and texts, while preserving everything else in the right second to-be-refined image untouched.''}}
As shown in the~\figref{fig:othermethod}, existing methods fail to perform consistent correction on generated images while maintaining background fidelity. In contrast, our method substantially improves image consistency while preserving the original background.

\section{Dataset Comparison}\label{dataset_comparison}
%  如图所示 我们展示了我们数据集中的pair 并且对比了我们数据集中的reference-target pairs和其他方法的成对数据 
% 从图中可以看出 我们的成对数据在包含多种语系 不同的场景 不同来源的生成方法和不同的视角的前提下 保证了数据的高一致性和局部的细节纹理 而其余数据集如Subject-200k 和UNO-1M 在局部细节上没有保证严格的一致性 或者出现局部模糊的情况 UNO-1M~\cite{uno} X2I2~\cite{omnigen2}, Subjects200K
As shown in the figure, we present several pairs from our dataset and compare our reference–target pairs with those from other datasets. From the visualizations, it is evident that our paired data maintain high global consistency and rich local details, even under diverse languages, scenes, generation sources, and viewpoints. In contrast, datasets such as Subjects-200k~\cite{ominicontrol} and UNO-1M~\cite{uno} do not strictly preserve local consistency and often exhibit region-level blurring or mismatches.

\section{Additional Visual Results }\label{visual}
\begin{figure*}
    \centering
    \includegraphics[width=1\linewidth]{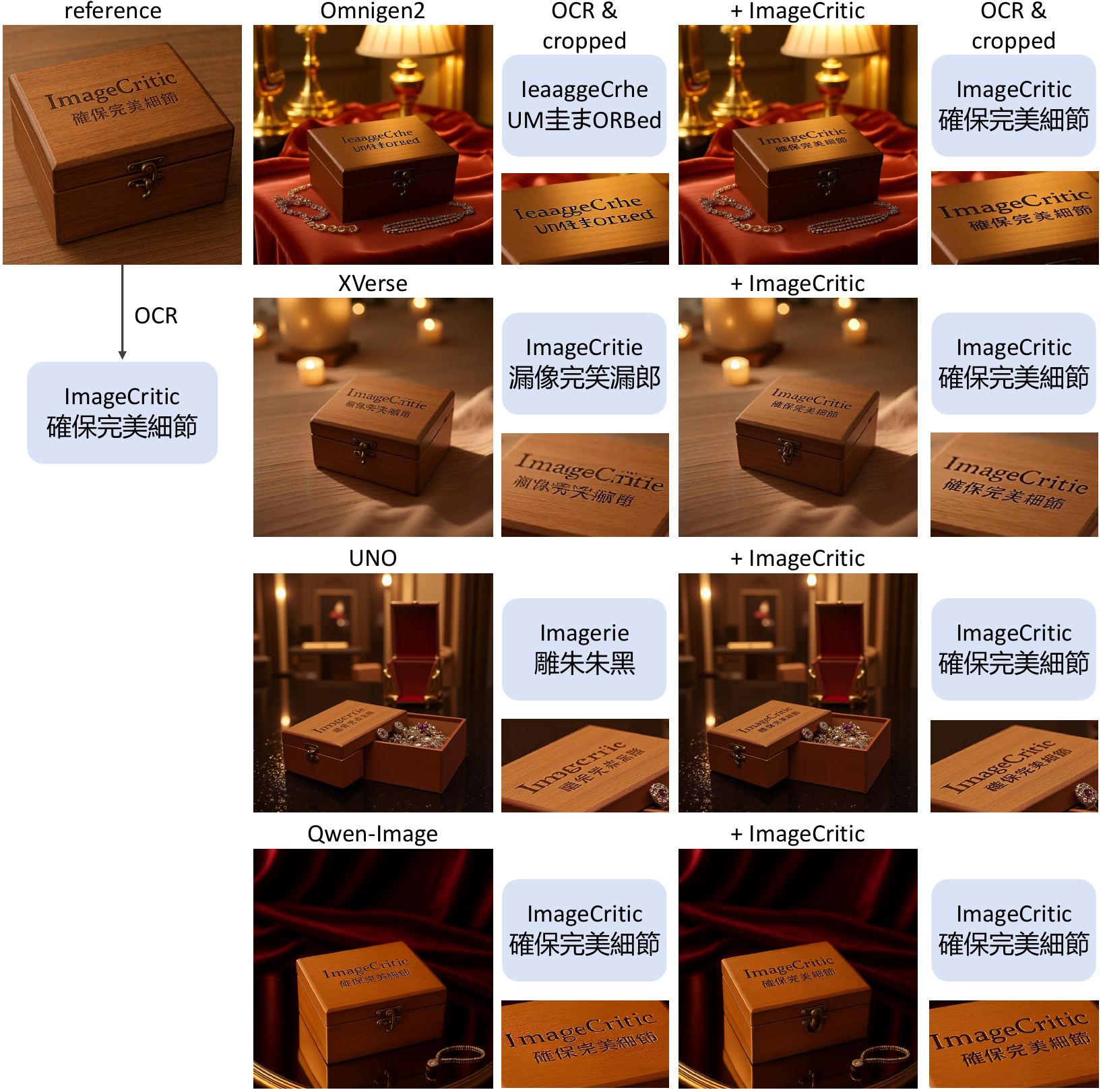}
% \textbf{addition visual result} 我们展示了我们的提出的方法在 多语言 多视角 多场景下的correcting结果 并且用ocr模型识别图中文字 可以看出使用我们方法后，生成图片的文字识别与参考图片之间全部一致 实现了对细节的完美correcting 
    \caption{\textbf{Additional visual result.} We present the visual results of our proposed \ourMthd{} under multilingual, multi-view, and multi-scene settings. By applying an OCR~\cite{ocr} model to the generated images, we observe that after correction with our method, all recognized text perfectly matches the reference images. This demonstrates that our approach achieves precise and comprehensive detail correction without disrupting the overall structural or contextual integrity of the images.}
    \label{fig:language}
\end{figure*}

\begin{figure*}
    \centering
    \includegraphics[width=1\linewidth]{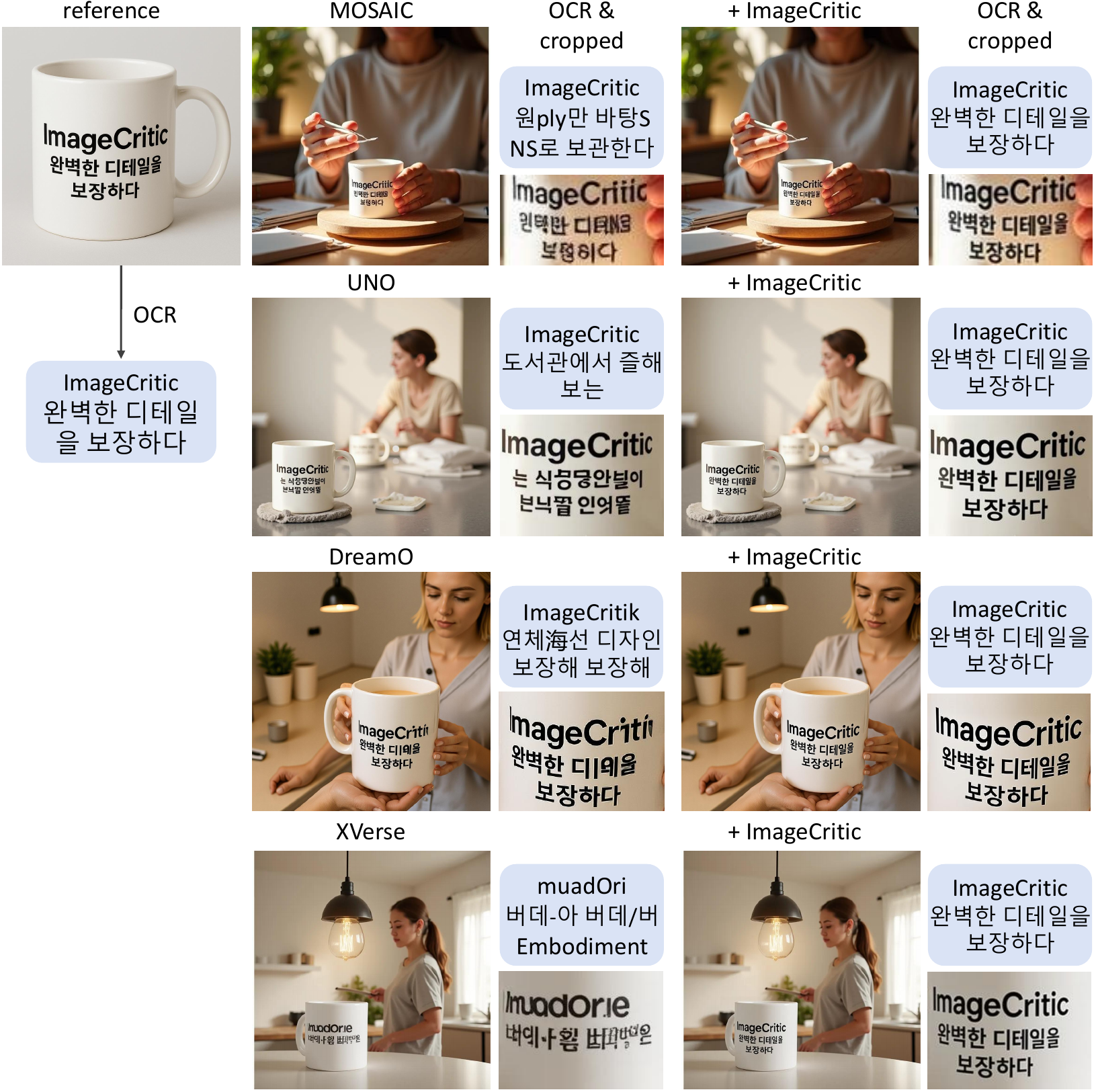}
    \caption{\textbf{Additional visual result.} We present the visual results of our proposed \ourMthd{} under multilingual, multi-view, and multi-scene settings. By applying an OCR~\cite{ocr} model to the generated images, we observe that after correction with our method, all recognized text perfectly matches the reference images. This demonstrates that our approach achieves precise and comprehensive detail correction without disrupting the overall structural or contextual integrity of the images.}
    \label{fig:language1}
\end{figure*}

\begin{figure*}
    \centering
    \includegraphics[width=1\linewidth]{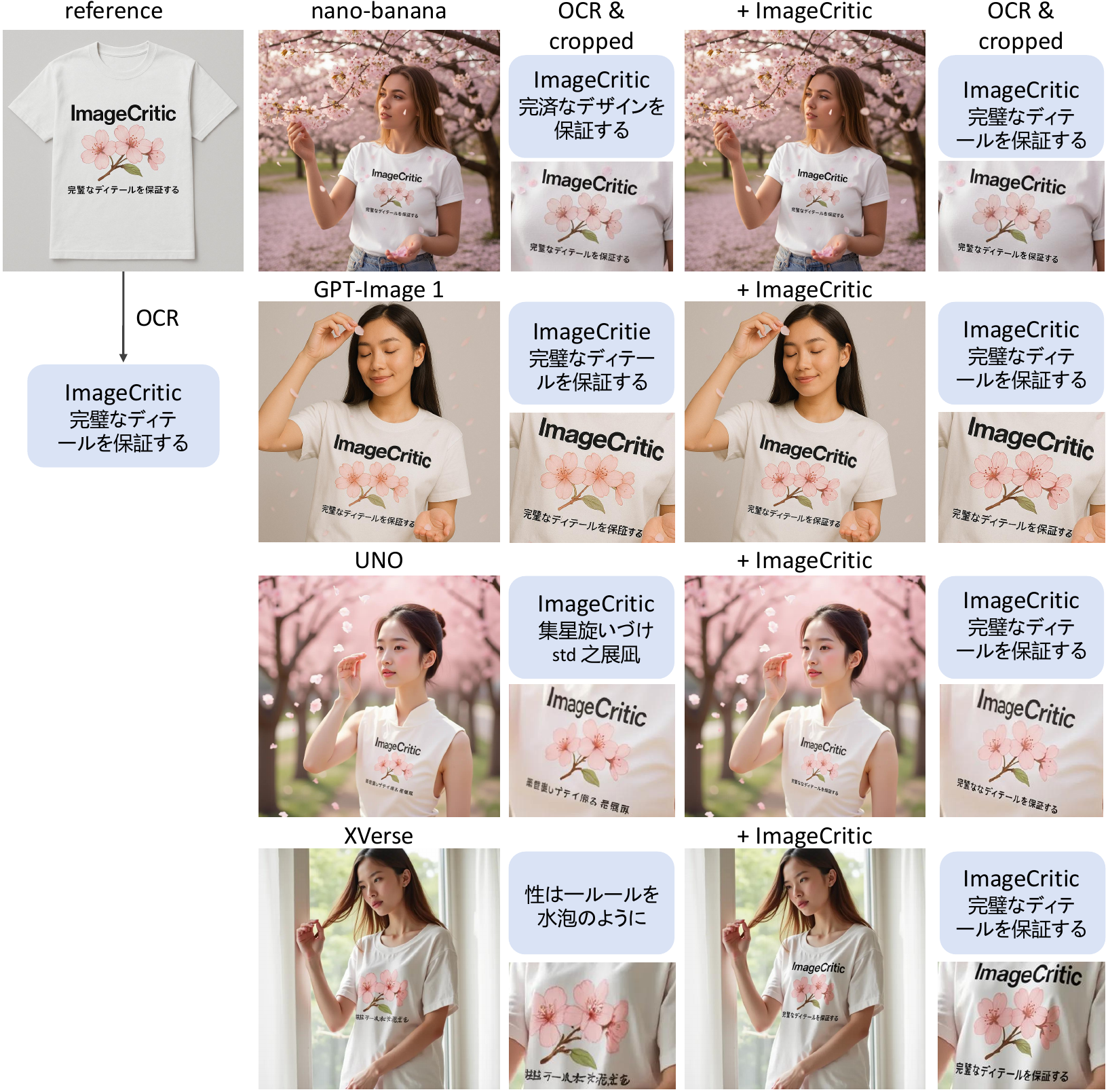}
    \caption{\textbf{Additional visual result.} We present the visual results of our proposed \ourMthd{} under multilingual, multi-view, and multi-scene settings. By applying an OCR~\cite{ocr} model to the generated images, we observe that after correction with our method, all recognized text perfectly matches the reference images. This demonstrates that our approach achieves precise and comprehensive detail correction without disrupting the overall structural or contextual integrity of the images.}
    \label{fig:language2}
\end{figure*}

% 为了进一步展现我们方法在不同语言 情景下的泛用性 我们首先使用gpt-image 1生成了不同语言的物品图片 随后使用了正文中提到pen-source models — XVerse~\cite{xverse}, Dreamo~\cite{dreamo}, MOSAIC~\cite{mosaic}, OmniGen2~\cite{wu2025omnigen2}, UNO~\cite{uno}, and Qwen-Image~\cite{qwenimage} — as well as the closed-source models NanoBanana~\cite{nanobanana} and GPT-Image~\cite{gpt} 进行客制化生成 并且使用我们的方法进行一致性修复 并且使用ocr模型检测生成部分的文字 如图~\figref{language} ~\figref{language1} ~\figref{language2} 所示 我们的方法成功对图片进行了一致性修复 ocr对文字的检测结果进一步证明了我们方法在不同场景 不用语言下的一致性修复能力
To further demonstrate the robustness of our method across different languages and scenarios, we first used GPT-Image 1 to generate images with diverse categories in multiple languages, and then performed customized generation using the open-source models XVerse~\cite{xverse}, Dreamo~\cite{dreamo}, MOSAIC~\cite{mosaic}, OmniGen2~\cite{wu2025omnigen2}, UNO~\cite{uno}, and Qwen-Image~\cite{qwenimage}, as well as the closed-source models NanoBanana~\cite{nanobanana} and GPT-Image~\cite{gpt}.
We then applied our method for consistency correction and used an OCR model~\cite{ocr} to detect the text in the generated images.
As shown in Figures~\figref{fig:language}, ~\figref{fig:language1}, and ~\figref{fig:language2}, our method delivers effective consistency restoration, and the OCR results further underscore its robustness across diverse linguistic and contextual settings.

\section{Agent Chain Details}\label{agent}
\begin{figure*}
    \centering
    \includegraphics[width=0.86\linewidth]{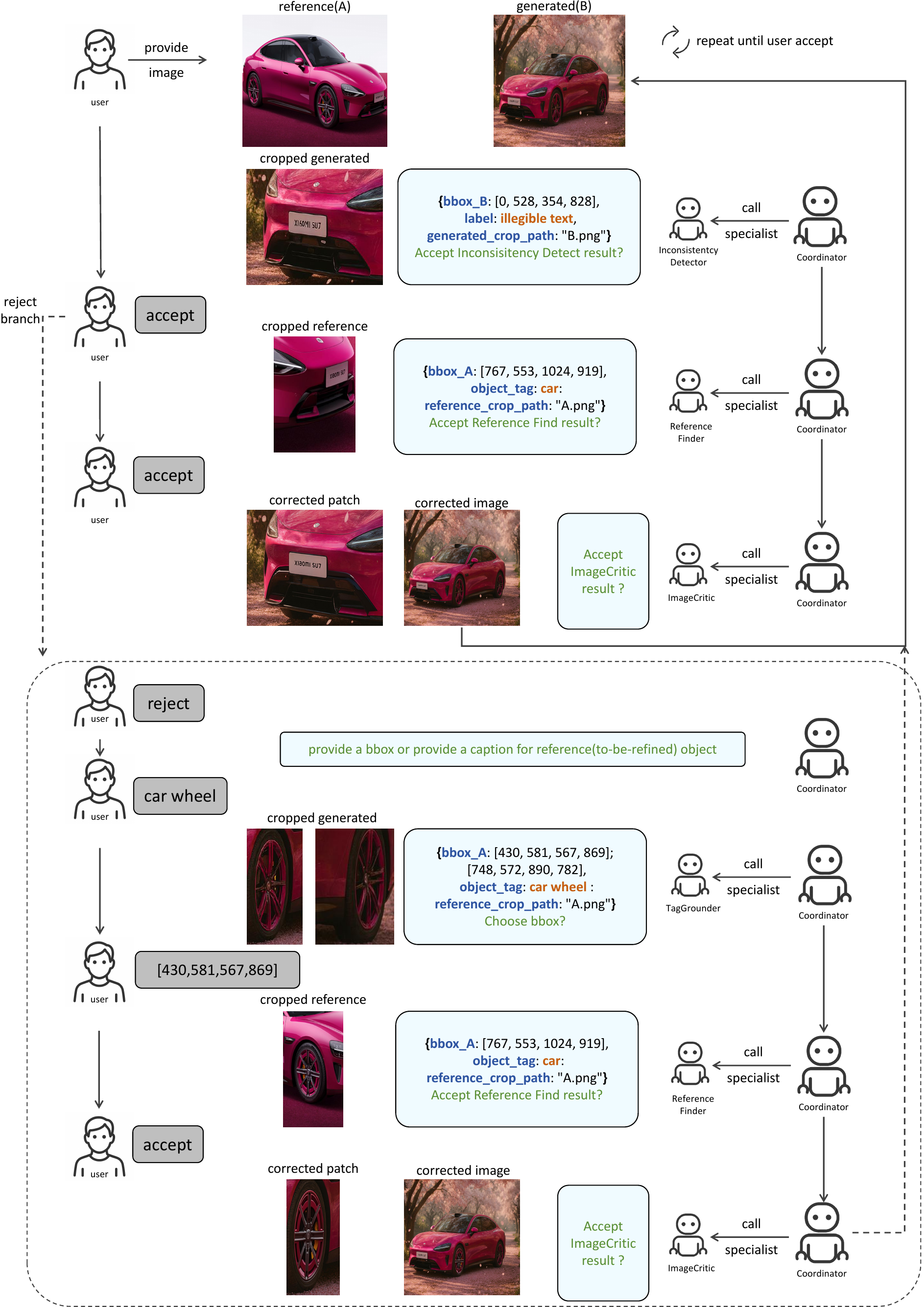}
    \vspace{-6pt}
    \caption{\textbf{Illustration of a multi-agent image correcting workflow.} 
    The system performs localized detection, reference matching, and iterative region-level corrections driven by user feedback, progressively correcting the generated image until it is accepted.}
    \label{fig:agent}
\end{figure*}

% 如图所示 系统通过多个专门代理的协同工作，对生成图像进行逐步改进。用户首先提供参考图像，随后不一致检测器、参考匹配器与图像评价代理依次对图像进行裁剪、比对与细化，直到用户接受结果。如果用户拒绝当前细化结果，则可以提供新的目标区域或文本描述，指导代理链进行进一步修正，该循环将持续进行，直至生成的图像满足用户意图。
As shown in Figure \ref{fig:agent}, the system progressively corrects the generated image through the coordinated operation of multiple specialized agents. After the user provides a reference image and a generated image, the inconsistency detector, reference finder, and ImageCritic agents sequentially perform comparison, cropping, and correction until the user is satisfied with the result. 
During this process, the user may choose whether to accept each proposed patch or provide a new target region or description to guide further correction by the agent chain. This iterative loop continues until the generated image aligns with the user’s intent. 
The prompts for the Inconsistency Detector, Reference Finder, TagGrounder, and Coordinator are configured in~\listref{lst:1}, ~\listref{lst:2}, ~\listref{lst:3}, and~\listref{lst:4}, respectively.
% 我们使用的

% figures

\begin{figure*}
    \centering
    \includegraphics[width=0.93\linewidth]{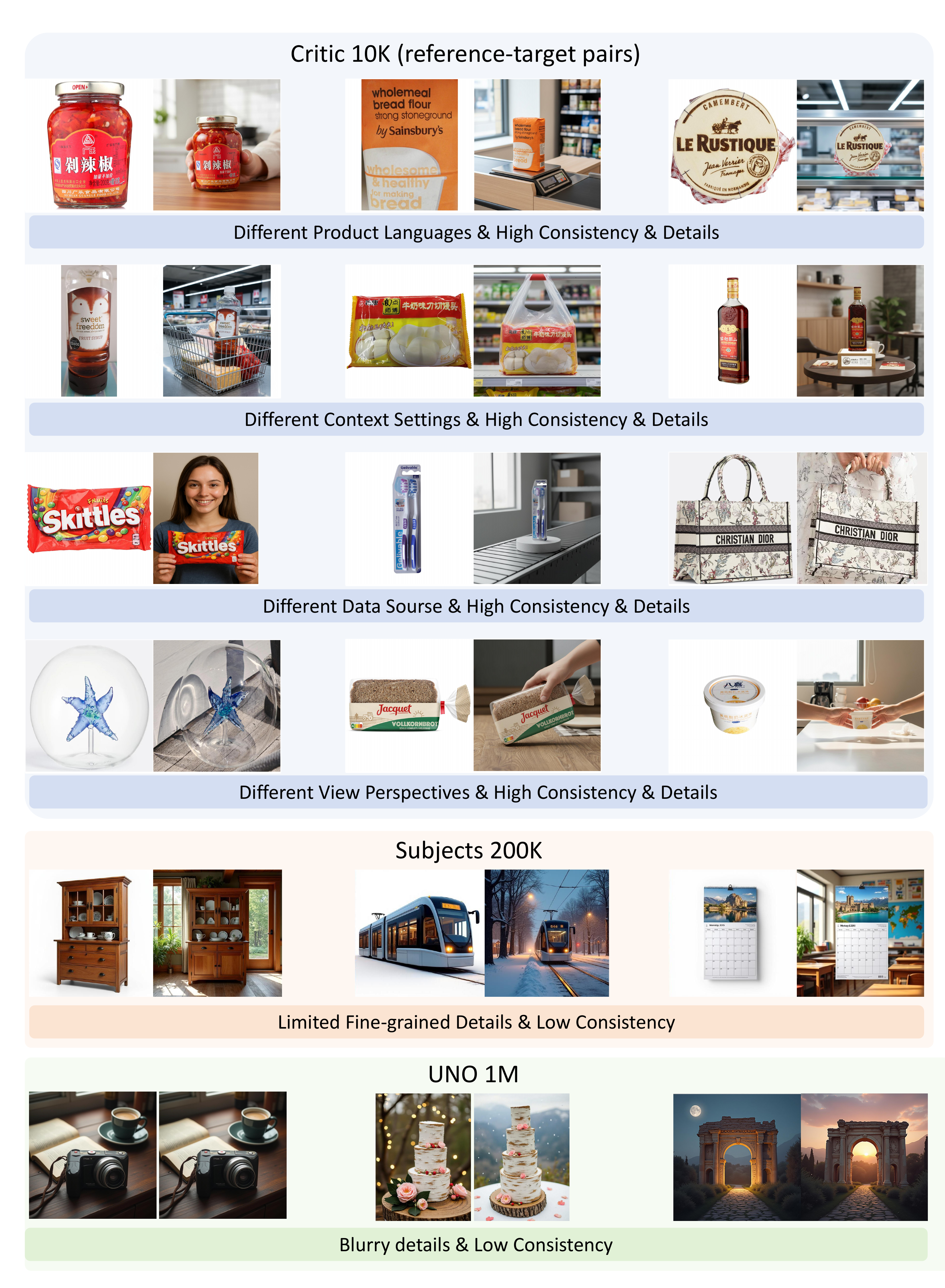}
    \vspace{-12pt}
    \caption{\textbf{Comparion with dataset.} Comparison of data samples from our Critic-10k dataset, the Subjects-200K~\cite{ominicontrol}, and the UNO-1M~\cite{uno} dataset.}
    
    \label{fig:placeholder}
\end{figure*}
\clearpage

\clearpage

\clearpage

\clearpage

\clearpage

% ###############################################
\begin{figure*}[t]
  \begin{lstlisting}[
    frame=tb,
    language=Python, 
    basicstyle=\ttfamily\small, 
    caption={The prompt for Inconsistency Detector. },
    label={lst:1}
  ]
    prompt = (
        "Carefully compare the two images. Image 1 is the reference image (correct version), and Image 2 is the target image that may contain defects. Focus only on the main subject of the image, ignoring any differences in the background. Identify the region in Image 2 that differs from the corresponding area in Image 1. Differences may include blur, illegible text, texture inconsistency, artifacts, missing parts, or any other visual discrepancies.Return ONLY the bounding box of the different region in Image 2 in the strict format:[xmin, ymin, xmax, ymax]"
    )

    messages = [
        {
            "role": "user",
            "content": [
                {"type": "text", "text": "Reference Image1 (correct version)"},
                {"type": "image", "image 1": image_A_path},
                {"type": "text", "text": "Target Image2 (may have defects)"},
                {"type": "image", "image 2": image_B_path},
                {"type": "text", "text": prompt},
            ],
        }
    ]
  \end{lstlisting}
\end{figure*}

\begin{figure*}[t]
  \begin{lstlisting}[
    frame=tb,
    language=Python,
    basicstyle=\ttfamily\small, 
    caption={The prompt for Reference Finder. },
    label={lst:2}
  ]
        prompt = (
            "I will show you a problematic region from image1 and a reference image2. "
            "Please find the corresponding region in image2 that matches the same area as the "
            "problematic region from image1. Return only the bounding box coordinates in the "
            "format [xmin, xmax, ymin, ymax], No additional text, just the coordinates inside []"
        )

        messages = [
            {
                "role": "user",
                "content": [
                    {"type": "text", "text": "image1 (problem region)"},
                    {"type": "image", "image": problem_crop_path},
                    {"type": "text", "text": "Reference image2"},
                    {"type": "image", "image": image_A_path},
                    {"type": "text", "text": prompt},
                ],
            }
        ]
  \end{lstlisting}
\end{figure*}
\clearpage

\begin{figure*}[t]
  \begin{lstlisting}[
    frame=tb,
    language=Python, 
    basicstyle=\ttfamily\small,
    caption={The prompt for TagGrounder. },
    label={lst:3}
  ]
        prompt = (
            f"Find the region in this image that best matches the product tag: \"{tag}\". "
            "Return ONLY the bounding box in Image in the strict format: "
            "[xmin, ymin, xmax, ymax]. No extra text."
        )

        messages = [
            {
                "role": "user",
                "content": [
                    {"type": "text", "text": f"Image to search product tag: {tag}"},
                    {"type": "image", "image": image_path},
                    {"type": "text", "text": prompt},
                ],
            }
        ]
  \end{lstlisting}
\end{figure*}
\clearpage

\begin{figure*}[t]
  \begin{lstlisting}[
    frame=tb,
    language=Python, 
    basicstyle=\ttfamily\small, 
    caption={The prompt for Coordinator. },
    label={lst:4}
  ]
        """
        You are the Coordinator Agent for an image restoration workflow.
        
        The workflow has three sequential steps:
        
        1. Inconsistency Detector: compare images and detect the difference region
           - Input: image_A (reference image path), image_B (target image path)
           - Output: bbox_B (difference region), prompt (problem description)
        
        2. Reference Finder: find a clean reference region
           - Input: image_A (reference), image_B (target), bbox_B (problem region)
           - Output: bbox_A (reference region), cropped reference region, object_tag
        
        3. ImageCritic: perform correction
           - Input: prompt, image_A, image_B, bbox_A, bbox_B, object_tag
           - Output: image_path (final corrected image), patch info
        
        Additionally, there is an auxiliary specialist:
        - TagGrounder: given an image and a user-provided product tag, locate a bbox in that image.
        
        Workflow rules:
        - Execute strictly in the order 1 - 2 - 3.
        - After each step, you MUST ask the user:
          "Accept [STEP_NAME] result? (yes/no):"
        - If the user answers "yes" or "y", continue to the next step.
        - If the user answers "no":
          - You must wait for the user to provide either:
            (a) a new bbox in the format [xmin, ymin, xmax, ymax], or
            (b) a product tag (e.g. 'shoe', 'bag', 'logo') to re-locate the region.
          - If the user gives a bbox, this bbox has the highest priority and must override previous bbox_B or bbox_A.
          - If the user gives a product tag, you MUST call the TagGrounder specialist via the `delegate_to_specialist` tool to convert this tag into a bbox, then use this bbox in the current or next step.
        - Always delegate concrete image-processing work to the appropriate specialist via the `delegate_to_specialist` tool.
        - Maintain data flow between steps (outputs from a previous step feed into the next).
        - When all three steps are done and the final restored image is produced, output:
          "Image restoration workflow completed!"
          and then stop.
        
        Current task:
        The user provides image_A and image_B paths. Please run the three-step correction workflow.
        """
  \end{lstlisting}
\end{figure*}

\clearpage

%%%%%%%%% REFERENCES

\end{document}